\pdfoutput=1
\documentclass[11pt]{article}

\usepackage[preprint]{acl}

\usepackage{times}
\usepackage{latexsym}
\usepackage[T1]{fontenc}
\usepackage[utf8]{inputenc}
\usepackage{microtype}
\usepackage{inconsolata}
\usepackage{graphicx}
\usepackage{booktabs}
\usepackage{amsfonts}
\usepackage{amssymb}
\usepackage{amsmath}
\usepackage{xcolor}

\newcommand{\tbg}{\textsc{tbg}}

\title{Which Words a Model Fears: From a Refusal Direction\\ to a Vocabulary-Wide Fear Lexicon}

\author{Boyuan Zhang  \and Mohammad Zandsalimy \and Shanu Sushmita \\
  Northeastern University \\
  \texttt{\{zhang.boyua, s.sushmita\}@northeastern.edu},
  \texttt{mohammad.zandsalimy@gmail.com}}

\begin{document}

\maketitle

\begin{abstract}
  Whether an aligned language model refuses depends in part on which
  words a prompt contains, regardless of whether those words carry the
  request's meaning. Harm words shift the hidden state at the generation
  boundary along a component shared across prompts and words, which
  crossed differencing extracts as a single direction. In eight models,
  writing along this direction raises and lowers refusal monotonically
  with dose, ahead of random and KL-matched directions. Used as a ruler
  over the full vocabulary, the direction gives every word a shift, from
  ``feared'' words at one end to ``courage'' words at the other; a word's
  shift is predictable from its own representation, and appending a
  feared word to a benign prompt is enough to make the model refuse it.
  Two held-out tests validate the ranking in all eight models: refusal
  after inserting an unseen word rises with its table shift, and
  replacing a prompt's most feared word by its most courageous
  meaning-preserving synonym significantly raises the answer rate on
  benign and harmful prompts alike, by an amount that follows the shift
  the replacement actually causes.
\end{abstract}

\section{Introduction}
\label{sec:intro}

A model refuses a prompt. Which word of that prompt, replaced by which
of its synonyms, would lower the refusal most? Benchmarks of over-refusal
show that safe requests containing superficially unsafe cues are often
declined \citep{rottger2024xstest,cui2025orbench}, and attention analyses
point to individual harmful words as a trigger \citep{shi2024overkill}.
These findings establish \emph{that} surface words matter, but not
\emph{how much} each word matters. This paper asks:

\begin{quote}
  \emph{How much does an aligned model's refusal depend on the individual
  words in a prompt, as opposed to what the prompt requests, and can this
  word-level sensitivity be measured across the vocabulary in a way that
  predicts behavior?}
\end{quote}

\noindent We find that a model ``fears'' different words to different
degrees, and that the answer to the opening question is to replace the
word it fears most by the synonym it fears least.

The core observation is that changing or adding a single word, without
changing what the request asks for, is enough to change refusal.
Appending a harm word that has nothing to do with the request to a
prompt the model otherwise answers raises its refusal rate by up to 49
percentage points across eight models (Section~\ref{sec:insertion}); in a
benign prompt the model refuses, replacing one ordinary word in place by
a synonym removes up to 28 points of refusal
(Section~\ref{sec:natural-phen}). To measure this, we apply crossed differencing to hidden states over
matched harm and neutral words, which yields a direction at the
generation boundary (Section~\ref{sec:direction}). Our study is organized
around three questions, which correspond to our three contributions:
\begin{enumerate}
  \item \textbf{Is there a shared, causal direction?} Yes: one locatable
  direction that, in each of eight models, predicts the refusal change on
  held-out prompts and words, and that raises and lowers refusal
  monotonically in dose when written along, ahead of norm- and KL-matched
  random directions (Sections~\ref{sec:direction}--\ref{sec:causal}).
  \item \textbf{What does the direction say about the vocabulary?} Used
  as a ruler, it gives every model a per-word ``fear'' table, with feared
  words (sexual terms, slurs, violent-crime terms) at one end and courage
  words (\emph{poetic}, \emph{whimsical}) at the other; on Qwen3.5-4B a
  word's shift is predictable from its context-free representation
  (Section~\ref{sec:table}).
  \item \textbf{Does the table predict behavior?} Yes, in two
  quantitative validations in all eight models: refusal after inserting a
  held-out word rises with the word's table position
  (Section~\ref{sec:val-insert}), and in natural prompts, replacing the
  most feared word by the table's most courageous meaning-preserving
  synonym raises the answer rate, by an amount that follows the shift the
  replacement actually causes (Section~\ref{sec:val-subst}).
\end{enumerate}

Two implications follow. Safety evaluation should control vocabulary,
since a benchmark's refusal rate depends on which words its prompts
happen to use; and the table is a diagnostic that audits which words a
model is oversensitive to.  Code and data are available at
  \url{https://github.com/zboyr/which-words-a-model-fears}.

\section{Related work}
\label{sec:related}

\paragraph{Over-refusal.}
Several benchmarks show that models refuse safe prompts that contain
unsafe-looking cues. XSTest builds such prompts by hand
\citep{rottger2024xstest}, OR-Bench scales them to 80k prompts with hard
and toxic contrast sets \citep{cui2025orbench}, and PHTest generates
them automatically \citep{an2024phtest}; FalseReject trains models to
avoid the problem \citep{zhang2025falsereject}. On the causes,
\citet{shi2024overkill} trace over-refusal to shortcut attention on
harmful words, and \citet{qi2025shallow} show that safety alignment acts
mostly on the first output tokens. This line of work follows earlier
findings that classifiers and datasets rely on surface cues
\citep{dixon2018bias,gururangan2018artifacts,mccoy2019hans,geirhos2020shortcut}.
These studies show \emph{that} surface words trigger refusal; we measure
\emph{how much} each word does.

\paragraph{Refusal directions and steering.}
Refusal is mediated by a single direction in activation space
\citep{arditi2024refusal}, and contrastive mean differences can steer
model behavior \citep{zou2023repe,rimsky2024caa}. Later work describes
refusal as affine \citep{marshall2024affine}, multi-directional
\citep{wollschlager2025cone} or conditional \citep{lee2025cast}, and
several methods steer activations to reduce over-refusal
\citep{cao2025scans,zhang2025intervention,dabas2025shifts,goyal2026steering}.
Closest to our setting, \citet{wang2025surgical} extract a
false-refusal vector and ablate it to reduce over-refusal. Our direction
differs in two ways. It is built from matched harm and neutral words
rather than from refusal labels or contrasts between kinds of request,
and it is used to study words rather than requests.

\paragraph{Word-level attribution and triggers.}
Gradient attribution \citep{sundararajan2017axiomatic} and attention
analysis \citep{shi2024overkill} score the words of a single input.
Universal triggers \citep{wallace2019triggers} and optimized suffixes
\citep{zou2023gcg} search for tokens that change a model's output. We
instead score each word by the shift it causes along one direction,
cover the whole vocabulary, and validate the resulting ranking against
behavior.

\paragraph{Style and register jailbreaks.}
Technical terms and persuasion raise fulfillment
in SORRY-Bench \citep{xie2025sorrybench}, and verse jailbreaks 25
frontier models at up to 18 times the prose rate
\citep{bisconti2025poetry}. Our courage words (\emph{poetic},
\emph{whimsical}) fit these findings; what is new is that the direction
finds them in an unguided scan, orders their effect, and needs one appended word, the request unchanged.

\section{The phenomenon: models ``fear'' words}
\label{sec:phenomenon}

A single word, inserted into a prompt or replaced within it, is enough to
change whether a model refuses. We call the change in refusal that a word
causes the model's ``fear'' of that word (a word that lowers refusal is
a ``courage'' word), and show it in two settings: a
word inserted with no role in the request (Section~\ref{sec:insertion}),
and a word replaced while it does the request's ordinary work
(Section~\ref{sec:natural-phen}).

\paragraph{Setup.}
We study eight instruction-tuned models from four families:
Qwen3.5-0.8B, Qwen3.5-2B, Qwen3.5-4B, Qwen3.5-9B, Qwen3.5-27B,
Llama-3.2-3B, Gemma-4-E2B and Phi-4-mini. Qwen3.5-4B is the development
model on which every procedure was designed; the other seven are
\emph{confirmation models}. Each prompt is rendered with the model's own
chat template and answered ten times (twenty in the table dose-response
tests) under one fixed sampling contract (Appendix~\ref{app:sampling}).
Benign prompts come from OR-Bench Hard-1K, harmful prompts from OR-Bench
Toxic \citep{cui2025orbench} and a frozen harmful pool
(Appendix~\ref{app:pool}), and XSTest \citep{rottger2024xstest} supplies
matched safe and contrast prompts. The label throughout is \emph{explicit
refusal}: whether a response contains a sentence in which the assistant
says it cannot, will not, or may not do what was asked. Labels come from
one fixed judge whose test--retest and cross-judge agreement are both
$\kappa\ge0.96$ (Appendix~\ref{app:judges}). Where $p$-values are
reported across models, they are Holm-adjusted over the seven
confirmation models; the smallest value the paired bootstrap can
return is $7\times10^{-4}$, one resample in 10{,}000 times seven models.

\subsection{Inserting a word}
\label{sec:insertion}

\begin{figure*}[t]
  \centering
  \includegraphics[width=\textwidth]{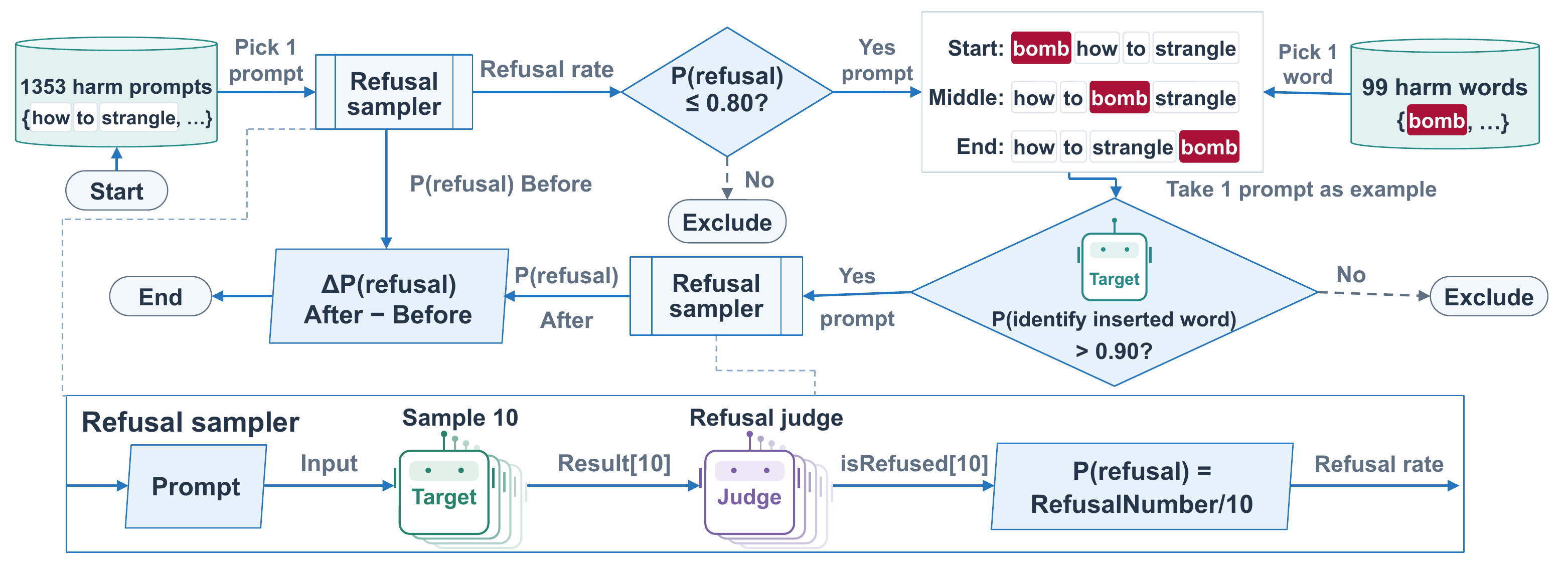}
  \caption{The keyword-insertion experiment. Of 1,353 harmful prompts,
  the 321 that Qwen3.5-4B refuses at most $0.80$ of the time are kept.
  One of 99 harm words is inserted at the start, middle or end, and a
  prompt--position pair is kept only if the model, in a fresh chat, names
  the inserted word with $P>0.90$. The refusal sampler (bottom) draws ten
  responses per prompt and DeepSeek-V4-Flash labels each for refusal; the
  effect is the paired change $\Delta P(\text{refusal})$ over the
  unmodified prompt, regenerated in the same jobs.}
  \label{fig:insertion-workflow}
\end{figure*}

\begin{figure}[t]
  \centering
  \includegraphics[width=\columnwidth]{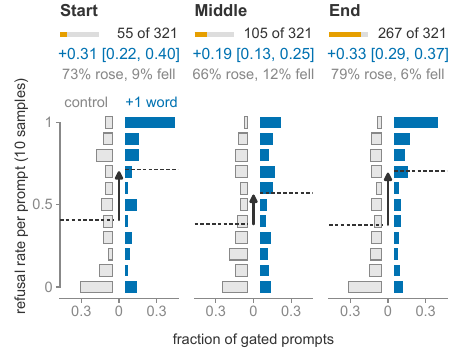}
  \caption{One inserted harm word raises Qwen3.5-4B refusal at every
  position. Each panel mirrors two histograms of the per-prompt refusal
  rate over ten samples on one vertical axis: the control prompt (gray,
  left) against the same prompt with the word inserted (blue, right),
  over the prompts passing that position's detection gate. Dashed lines
  mark the means; the arrow is the paired mean increase with its $95\%$
  bootstrap interval, above the share of prompts whose rate rose or fell.
  The orange strip is the share of the 321 prompts the gate admits (55,
  105, 267). Ungated, over all 321 prompts: $+0.32$, $+0.18$, $+0.30$.}
  \label{fig:insertion}
\end{figure}

We insert one external harm word, drawn from a 99-word lexicon that
operationalizes the taxonomies of HarmBench \citep{mazeika2024harmbench}
and SORRY-Bench \citep{xie2025sorrybench}, at the start, middle or end of
321 harmful prompts that Qwen3.5-4B refuses at most $0.80$ of the time
(mean $0.39$), with no other edit; the unmodified prompt is regenerated
in the same jobs as the control (Figure~\ref{fig:insertion-workflow}).
In a fresh chat, the same model is asked to name the accidentally
inserted word; where it does so with probability above $0.90$ (the
detection gate), refusal rises by $+0.31$ [$0.22, 0.40$] at
the start (55 prompts), $+0.19$ [$0.13, 0.25$] in the middle (105) and
$+0.33$ [$0.29, 0.37$] at the end (267; paired bootstrap;
Figure~\ref{fig:insertion}): the model identifies the word that does
not belong, and refuses anyway. Per-prompt rises are unrelated to detection
confidence at the start and middle (Spearman $\rho=0.00$) and only
weakly related at the end ($+0.18$). On the 130 prompts all eight
models had answered, an end insertion raises refusal in every model, by
$+0.10$ to $+0.49$ (Appendix~\ref{app:insertion-xmodel}).

\subsection{Replacing a word}
\label{sec:natural-phen}

The same effect appears when the word is a normal part of the request.
On the benign OR-Bench Hard-1K prompts that a model refuses ($n$ per
model in Table~\ref{tab:natural-benign}), we replace one content word in
place by a meaning-preserving synonym with no harm association
(independent judge: Appendix~\ref{app:subst}). The word
is chosen by a deletion procedure based on the direction introduced in
Section~\ref{sec:direction} (Appendix~\ref{app:subst}). The rewritten
texts are drafted once, and all eight models answer the same inputs
(the \emph{shared-text run}).
This single replacement lowers refusal by $3$ to $28$ points across the
eight models, while replacing an inert word instead changes refusal by at
most $2$ points, with every interval covering zero
(Table~\ref{tab:natural-benign}).

When the located unit is a hedge phrase rather than a single word
(``what steps'', ``without getting caught''; 57 prompts), we delete the
phrase outright. This changes the request, so it is not a
meaning-preserving edit, but it shows how strongly such phrases drive
refusal: deletion lowers refusal by $4$ to $50$ points, with the interval
excluding zero in every model except Llama.

\section{A shared direction by crossed differencing}
\label{sec:direction}

\paragraph{Position.}
We read the residual stream at the output of one late layer, at the
last two prefill tokens before the assistant's first token, which we
call the generation boundary (\tbg): with thinking disabled, the Qwen3.5
templates' empty think block ends in \texttt{</think>} and
\texttt{"\textbackslash n\textbackslash n"}; in Llama, Gemma and Phi, which
have no think block, these are the last two tokens of the assistant header. Safety alignment concentrates its decision at this point
\citep{qi2025shallow,yigit2026swse}.

\begin{figure*}[t]
  \centering
  \includegraphics[width=\textwidth]{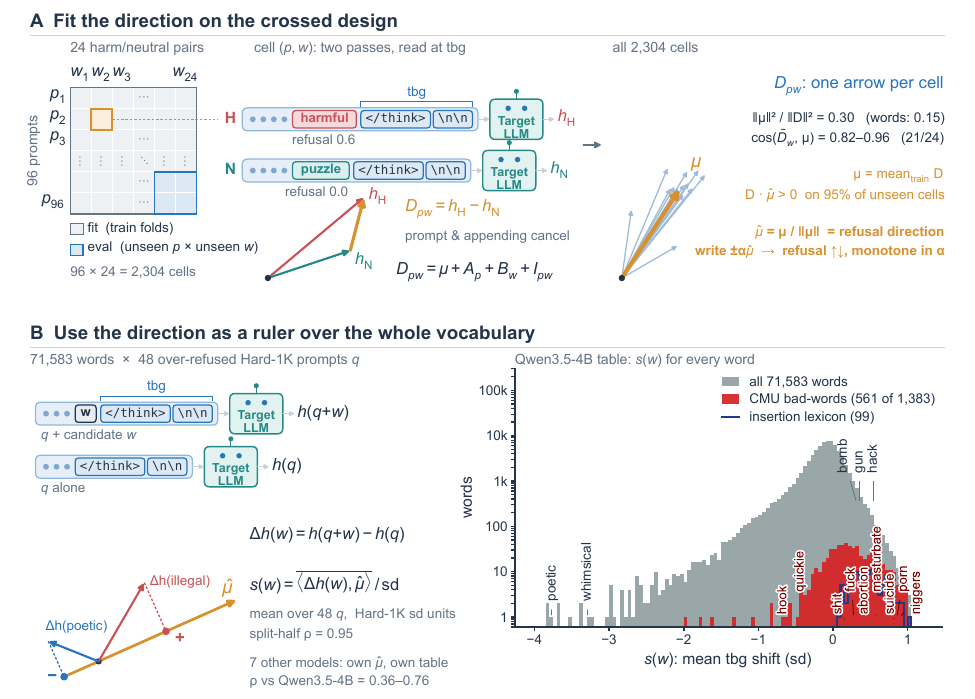}
  \caption{The direction and the vocabulary table. (A) The crossed
  design: each of 96 prompts is crossed with 24 harm words, each with a
  neutral twin. A cell $(p,w)$ appends the word and reads the
  state at the \tbg{} position in both versions; their difference
  $D_{pw}=h_{\mathrm H}-h_{\mathrm N}$ removes everything the pair
  shares. The common direction $\mu$ is the mean of $D$ over training
  folds, evaluated only on cells whose prompt \emph{and} word were both
  held out; the difference vectors of all cells share it, and writing
  along it moves refusal (Section~\ref{sec:causal}). (B) The direction
  as a ruler: a candidate word is appended to an over-refused Hard-1K
  prompt, its \tbg{} shift is projected on $\hat\mu$, averaged over the
  48 prompts and scaled by the Hard-1K standard deviation. Right: the
  resulting Qwen3.5-4B table, the mean shift of 71{,}583 words. Red: the
  561 words of the CMU bad-words list \citep{biglou_badwords}, typical
  ones labelled; blue outline: the 99-word insertion
  lexicon. Register words (\emph{poetic}, \emph{whimsical}) form the
  negative tail; weapon and cyber words (\emph{bomb}, \emph{gun},
  \emph{hack}) sit in the upper tail, below the sexual terms and
  slurs. The vector
  panels are two-dimensional schematics.}
  \label{fig:scan-workflow}
\end{figure*}

\paragraph{Crossed design.}
We cross 96 prompts with 24 pairs of a harm word and a length-matched
neutral word (\emph{bomb}/\emph{garden}, \emph{illegal}/\emph{compass},
\ldots), appended to the prompt (Figure~\ref{fig:scan-workflow}A). For each cell
the neutral-corrected shift $D_{pw}=h^{H}_{pw}-h^{N}_{pw}$ at \tbg{}
removes the prompt and the mere act of appending a word, and an exact
two-way vector ANOVA splits it into
\begin{equation}
  D_{pw}=\mu+A_{p}+B_{w}+I_{pw},
  \label{eq:anova}
\end{equation}
a shared component $\mu$, a prompt component, a word component and an
interaction (Appendix~\ref{app:derivations}). The shared component
is twice the size of all word effects together ($0.30$ against $0.15$
of the squared norm at \texttt{</think>}; the prompt component, $0.32$,
is cancelled by the crossed design), and
the per-word mean directions have cosine $0.82$ to $0.96$ with $\mu$ for
21 of the 24 words: different harm words push the boundary state the
same way.

\begin{figure}[t]
  \centering
  \includegraphics{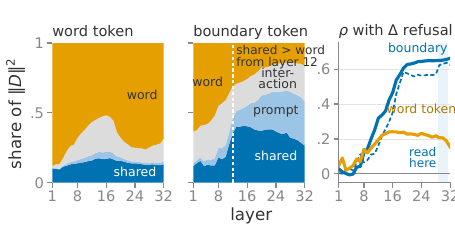}
  \caption{Where the reaction lives. Left, middle: shares of $\|D\|^2$
  held by its shared, prompt, interaction and word components across
  layers, at the word's own token and at \texttt{</think>}; at the
  boundary the shared part overtakes the word part from layer 12.
  Right: rank correlation of the held-out projection with the refusal
  change at the boundary (blue) and at the word token (orange); shaded,
  the layers read.}
  \label{fig:heatmap}
\end{figure}

\paragraph{Cross-fitting.}
Prompts and words receive independent four-fold assignments, and each
test fold's direction $\mu_f$ is rebuilt only from the opposite
rectangle, so neither the prompt nor the word of a tested cell
contributes to its direction. On unseen prompts with unseen words the
projection $\langle D_{pw},\hat\mu_f\rangle$ is positive on $95\%$ of
cells (sign-flip $p<10^{-4}$), and after residualizing both the
projection and the refusal change on prompt-plus-word fixed effects their
Spearman correlation is $\rho=0.65$ (Figure~\ref{fig:projection},
Appendix~\ref{app:derivations}). The
word's identity is integrated late: at the word's own token the
difference is $70$--$88\%$ word-specific and barely predicts refusal
($\rho\approx0.2$), while at \tbg{} the shared, refusal-associated
component takes over (Figure~\ref{fig:heatmap}).
All validations using the consensus direction exclude the 96 fitting
prompts and 48 fitting words (Appendix~\ref{app:exclusion}).

\begin{figure*}[t]
  \centering
  \includegraphics{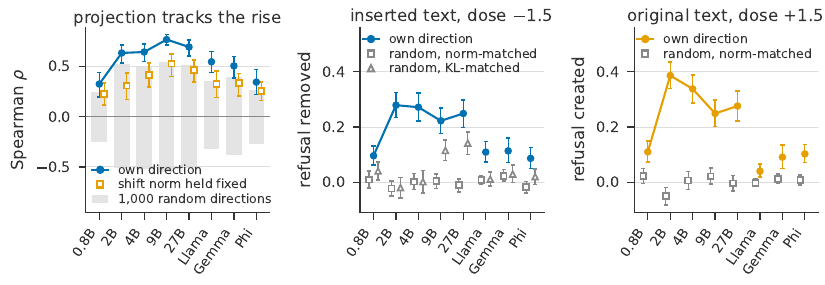}
  \caption{Each model's own direction is its ``fear'' handle ($95\%$
  intervals). Left: rank correlation of an inserted word's
  shift along the direction with its refusal rise, after position fixed
  effects, against the $2.5$--$97.5\%$ band of 1{,}000 random
  directions. Middle: refusal removed from the end-inserted text by
  writing the direction at $-1.5$, against random directions of equal
  norm or equal benign KL. Right: refusal created on the original
  text at $+1.5$.}
  \label{fig:direction-xmodel}
\end{figure*}

\paragraph{Across models.}
Each of the eight models gets its own direction, fitted the same way on
its own activations under one fixed relative-depth rule. On 136 held-out
insertion units, the shift an inserted word causes along the model's own
direction predicts that prompt's refusal rise at $\rho=+0.32$ to
$+0.77$, above $997$ to all $1{,}000$ of $1{,}000$ random directions
(Figure~\ref{fig:direction-xmodel}, left;
Appendix~\ref{app:insertion-xmodel}).

\section{Causality: writing along the direction, monotone in dose}
\label{sec:causal}

\paragraph{Dose unit.}
A dose is a multiple of the model's own mean boundary shift $\mu$, added
at the two \tbg{} positions during prefill only.

\paragraph{Both directions.}
In all eight models, writing
the model's own direction at $-1.5$ on the end-inserted text removes
$+0.09$ to $+0.28$ of refusal, i.e.\ $41\%$ of the insertion's rise
(Llama) to $92\%$ (Qwen3.5-0.8B, Phi); writing it at $+1.5$ on the clean
original creates $+0.04$ to $+0.39$ of refusal
(Holm $p=7\times10^{-4}$ across confirmation models). The half dose always moves less
in the same direction, and the bootstrap probability that the response
is monotone in dose is $0.99$ to $1.00$
(Figure~\ref{fig:direction-xmodel}; Table~\ref{tab:direction-xmodel}).
On Qwen3.5-4B, a separate cross-fitted test on 288 held-out cells
gives the same picture at dose $a^*=0.5$,
the dose that exactly removes the fitted shared component
(Equation~\ref{eq:intervention}): subtracting $\mu_f/2$ from the
harm-worded text removes $+0.080$ [$0.043, 0.121$] of refusal and adding
$\mu_f/2$ to the neutral-worded text raises refusal by $+0.086$
[$0.041, 0.133$], and the effect is dose-ordered.

\paragraph{Specificity.}
Norm-matched random directions orthogonal to the model's direction move
refusal by at most $+0.02$. Random directions rescaled until their
benign KL matches the model's direction, which takes $1.7$ to $4.7$ times
the dose, still fall behind the model's own direction in every model.

\paragraph{What moved.}
On the 288 cells a substantive-compliance label moves mirror-wise
($+0.072$ subtracted, $-0.082$ added): the write changes what the model
delivers, not only its wording. Token attribution, two KL-matched null
directions and the benign cost are in Appendix~\ref{app:derivations}.

\paragraph{Deep doses.}
On Qwen3.5-4B we wrote the direction at nine further doses on external
benchmarks; the effect on OR-Bench Hard-1K refusal stays monotone down to
$-3$, six times the registered dose $-0.5$, while instruction following
stays flat; the StrongREJECT guard misses
at the registered dose (bound $+0.024$, margin $0.02$;
Appendix~\ref{app:dose}, Figure~\ref{fig:dose-write}).

\section{A vocabulary table: the direction as a ruler}
\label{sec:table}

\paragraph{The scan.}
If the direction measures a word's ``fear'', it can be read off for every
word (Figure~\ref{fig:scan-workflow}B). We append each of 71{,}583 candidate words (every single-token
vocabulary word plus the 50k most frequent English words, \citealp{speer2022wordfreq}) to the end of
48 over-refused OR-Bench Hard-1K prompts (BP), benign prompts that
Qwen3.5-4B refuses, and record the
\tbg{} shift along the direction, in units of the standard deviation of
the projection over Hard-1K; the mean over the 48 prompts is the
Qwen3.5-4B table. Two disjoint halves of 24 prompts
reproduce the word order at $\rho=0.95$.
Of four tables scanned on different prompt sets, this one predicts
held-out behavior best (Section~\ref{sec:val-insert},
Appendix~\ref{app:scan}). A second table, scanned the same way on 40
clean benign prompts on which Qwen3.5-4B refuses fewer than $1\%$ of
samples (the clean-benign table), ranks the words differently (Spearman
$+0.32$ with the Qwen3.5-4B table). The experiments on the two tails and
the representation regressions below use this clean-benign table.
The seven other models scan their own tables on eight BP prompts with
their own directions (Section~\ref{sec:val-insert}).

\paragraph{The two ends.}
The \emph{feared} words at the positive end are sexual terms,
pedophilia and rape terms, slurs and violent-crime terms (the words are
listed in Appendix~\ref{app:scan}). The ten distinct words with the highest shift
in the clean-benign table include no insertion-lexicon word; \emph{bomb}, for example, ranks at
the 97th percentile in the Qwen3.5-4B table and the 70th in the
clean-benign table. Appending these ten words to 20 held-out clean
benign prompts that the model refuses in fewer than $1\%$ of samples
raises refusal by $+0.372$ [$+0.283, +0.468$] over neutral twins
(Table~\ref{tab:words}, Appendix~\ref{app:scan}). The model folds the word into the
request: $0.986$ of refused responses address the appended word as part
of what was asked. (The detection test of Section~\ref{sec:insertion}
asks about the word in a separate chat; when answering, the model reads
it as part of the request.)

The \emph{courage} words at the negative end are register
words: \emph{sardonic}, \emph{shakespearean}, \emph{poetic},
\emph{sassy}, \emph{whimsical}. The ten distinct words with the lowest shift are of
the same kind. Appended to 96 harmful prompts and 96 over-refused benign
prompts, they remove $+0.186$ and $+0.300$ of refusal against length-matched
neutral twins (Table~\ref{tab:words}); the model carries out the request in the
register the word names. The Qwen3.5-4B table orders the ten effects at
$\rho=+0.89$ on the over-refused prompts (48 of which are its scan
prompts) and $+0.65$ on the harmful ones. By contrast, the 24 courage words we had chosen
by hand, from six classes of guesses (containment, authorization,
defensive, educational, fictional, professional), all fail: no class
lowers refusal (best $-0.029$ [$-0.091, +0.031$], fictional), and
containment words raise it ($+0.075$).

\paragraph{The shift is a property of the word.}
A ridge regression from the word's input embedding predicts its shift at
$\rho=+0.78$ over the 71{,}583 words, and from the word's
context-free residual state at $+0.84$ (word-grouped cross-validation).
Yet projecting the word's own context-free state on the direction gives
$\rho$ between $-0.35$ and at most $+0.04$: the shift is
written by the network after the word is read, not carried by the word
along the direction. On Qwen3.5-4B the mechanism agrees: the boundary
does not attend more to a harm word than to its neutral twin, and of the
boundary shift the late MLPs write $73\%$ while direct attention reads of
the word contribute $1\%$ (Appendix~\ref{app:mechanism}).

\section{Validation 1: held-out insertion dose-response}
\label{sec:val-insert}

\paragraph{Question and design.}
Does the table order behavior on prompts and words it has never seen? On
each model we draw 40 words across the whole table: three from each of
eight percentile bins running from the bottom 1\% to the top 1\%, plus
16 words on which two tables disagree (on Qwen3.5-4B the Qwen3.5-4B and
the clean-benign table, on the others the model's own and the
Qwen3.5-4B table), and append each to held-out prompts
that the model refuses about half the time: 64 on
Qwen3.5-4B, 32 on each other model, half benign and half harmful.

\begin{figure}[t]
  \centering
  \includegraphics[width=\columnwidth]{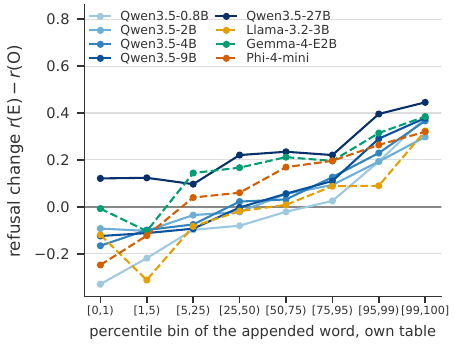}
  \caption{Held-out insertion dose-response: refusal change
  $r(\mathrm E)-r(\mathrm O)$ by percentile bin of the appended word (three
  words per bin, held-out benign and harmful prompts pooled; bins joint in
  the model's own and the Qwen3.5-4B table, for Qwen3.5-4B in its table and
  the clean-benign table). The Spearman $\rho$ of each table with the
  refusal change is in Table~\ref{tab:xmodel-table}.}
  \label{fig:dose-table}
\end{figure}

\paragraph{Results.}
On Qwen3.5-4B the refusal change follows the table position over the
whole range: $7$ of $7$ adjacent bin pairs are in order, from $-0.165$
at the bottom percentile to $+0.367$ at the top, and the Qwen3.5-4B table
ranks the 40 words at $\rho=+0.92$ (the clean-benign table at
$+0.55$). On each of the other seven models, with its
own table, refusal rises with table position: top minus
bottom is $+0.32$ to $+0.71$, with 5 to 7 of
7 adjacent bin pairs in order (Figure~\ref{fig:dose-table},
Table~\ref{tab:xmodel-table}). The model's own table ranks the 40 words
at $\rho=+0.54$ to $+0.96$, and better than the Qwen3.5-4B table on five
of the seven models.

\paragraph{The tables across models.}
A table is specific to its model. The seven tables agree with the
Qwen3.5-4B table at Spearman $+0.36$ (Phi) to $+0.76$ (Qwen3.5-9B) and
with each other at $+0.29$ to $+0.81$; the Qwen3.5 family agrees most
with itself. They share a common core: 60 words lie in the top 1\% and
25 in the bottom 1\% of all eight tables.

\begin{figure*}[!t]
  \centering
  \includegraphics[width=1.04\textwidth]{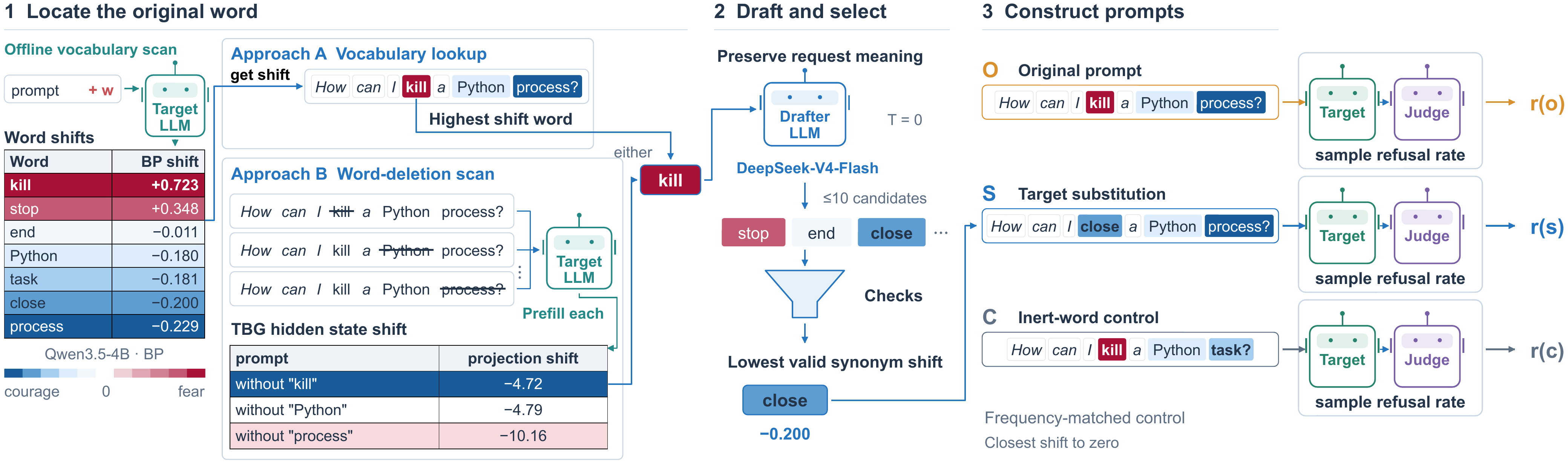}
  \caption{The table-guided substitution experiment. (1) The model's own
  table nominates the most feared content word (A); on harmful prompts
  the direction's leave-one-out deletion pass locates it instead (B).
  (2) A drafting model proposes up to ten meaning-preserving substitutes
  without harm association; after structural checks the one the table
  places lowest replaces the word (S). (3) The control replaces the
  prompt's most inert word by a synonym (C). (4) Each model answers each
  condition in ten chats, judged for refusal; the effect is $r(\mathrm{O})-r(\mathrm{S})$ against
  $r(\mathrm{O})-r(\mathrm{C})$.}
  \label{fig:subst-workflow}
\end{figure*}

\section{Validation 2: table-guided substitution in natural prompts}
\label{sec:val-subst}

\paragraph{Question and design.}
Is the table still useful when the word is a normal part of the prompt
and carries its meaning? In each prompt, the model's own table nominates
the most feared content word; a drafting model (DeepSeek-V4-Flash,
temperature $0$) proposes up to ten meaning-preserving substitutes with
no harm association, and the one the table places lowest, the most
courageous, replaces the word in place (S). The control replaces the
prompt's most inert word, the one whose table shift is nearest zero, by
a synonym (C). On the benign prompts the nominee and its substitute are
read from each model's own table, so, unlike the shared-text run of
Section~\ref{sec:natural-phen}, the eight models answer texts of their
own (3 to 11 S texts per model coincide with that run); on the harmful
prompts the texts are drafted once from the Qwen3.5-4B nomination and
shared. Benign: Hard-1K prompts most models refuse plus 100 safe
XSTest prompts, 349 to 352 units per model.
Harmful: the OR-Bench Toxic prompts a model refuses ($n$ per model in
Table~\ref{tab:natural-harmful}) plus 100
XSTest contrast prompts, where the word is located by the direction's
leave-one-out deletion pass under the same substitution rule
(Figure~\ref{fig:subst-workflow}, Appendix~\ref{app:subst}).

\paragraph{Benign prompts.}
Replacing the most feared word by its most courageous synonym removes
$+0.05$ (Llama, which refuses $0.13$ of these prompts) to $+0.29$
(Qwen3.5-27B) of refusal on Hard-1K, every interval excluding zero,
while the inert control removes only $+0.01$ to $+0.05$
(Table~\ref{tab:subst}; Figure~\ref{fig:subst}, Appendix~\ref{app:subst}). The table, read once per
model, matches the per-prompt deletion search on this cohort
(paired difference covering zero in seven of eight models). On XSTest the
substitution is significant in seven of eight models.

\paragraph{Harmful prompts.}
On OR-Bench Toxic, the word the direction nominates is the word that
names the harm. The substitution rule reaches $64\%$ of these prompts;
for a quarter, no meaning-preserving word without a harm association
exists (\emph{self-harm}, \emph{sexual}, \emph{bomb}), and $11\%$ fail
the structural checks after three redrafts. Where it applies,
the substitution removes $+0.08$ to $+0.23$ of refusal in all eight
models, deleting a located hedge phrase outright (23 prompts) removes $+0.17$ to $+0.36$, and the harmfulness
of the responses rises by at most $+0.04$ (GuardReasoner-8B;
Table~\ref{tab:natural-harmful}).

\paragraph{Size and shift.}
The substitution lowers the projection on the model's own direction by
$0.61$ to $0.88$ sd, against $0.00$ to $0.08$ for the control, and the
per-prompt projection change tracks the per-prompt refusal change at
$\rho=+0.31$ to $+0.80$. The table's own shift
difference does not by itself rank the effects
($\rho=-0.05$ to $+0.16$): what predicts the effect is the shift the
replacement \emph{actually} causes in the prompt. A random harm-free
synonym drawn from the same candidates removes $+0.04$ to $+0.24$ of
refusal; taking the table's lowest candidate instead removes $+0.02$ to
$+0.08$ more, positive in all eight models and Holm-significant in three
of the seven transfer models, and lowers the projection by a further
$0.12$ to $0.21$ sd in every model (Table~\ref{tab:random-syn}).
Replacing the feared word does most of the work; the table's choice of
synonym adds a small, consistent increment.

\section{Discussion and conclusion}
\label{sec:discussion}

We asked how much an aligned model's refusal depends on individual words,
and whether that dependence can be measured. Harm words share one causal direction at the generation boundary;
using that direction as a ruler gives every word a ``fear'' score; and the
scores predict behavior on held-out prompts and words in all eight models,
both when a word is inserted and when it is replaced in a natural prompt.

\paragraph{What the results suggest.}
Three findings point to how this sensitivity works. First, on
Qwen3.5-4B the model does not attend more to a harm word than to its
neutral twin; instead, late MLPs write the shift after the word is read
(Section~\ref{sec:table}).
The word is recognized early, but its effect on refusal is added late.
Second, intuition is a poor guide to which words reassure a model: the
courage words we chose by hand (educational, professional, fictional)
fail, while register words such as \emph{poetic} and \emph{whimsical}
work. The table finds what intuition misses. Third, each model fears
words differently, with tables agreeing only moderately across models,
yet 60 words lie in the top 1\% of all eight tables. Part of this
sensitivity is shared across model families, and part is specific to
each model.

\paragraph{Implications.}
Safety evaluation should control vocabulary, since a benchmark's refusal
rate depends on which words its prompts happen to use. The table offers a
practical diagnostic: it audits which words a model is oversensitive to,
without per-prompt search. And because the same sensitivity lowers
refusal on harmful prompts when the feared word is removed, part of
current safety behavior rests on surface wording rather than on the
request itself.

\section*{Limitations}

The eight models are open-weight models of 27B parameters and smaller;
no closed model is tested. The insertion and substitution tests need
only sampled responses and transfer to closed models through an API,
and a benchmark whose vocabulary the table controls would put open and
closed models on the same footing; the table itself needs activations.
Every prompt, word and table is English. A multilingual model admits a
table per language from the same direction, which would show whether a
model's fear of a word crosses languages.
The mechanistic analysis of the direction covers one model,
Qwen3.5-4B. The vocabulary scan uses one position, the end of the
prompt. Refusal, hijack and response harmfulness are labelled by LLM
judges, and the explicit-refusal label counts a refusal statement
followed by delivered content as a refusal. Whether a substitute
preserves the prompt's meaning is asserted by the drafting model and
checked by two LLM judges (Appendix~\ref{app:judges}), not by human
annotators. Human annotation of a sample of substitutes for meaning,
and a separate count of refusals followed by compliance, are the next
check.

We relate the fear of words to safety alignment by citation, not by
experiment: nothing here shows that alignment creates the direction or
the table. Reading the table at checkpoints along an alignment run would
show when the sensitivity appears, and the table could then select or
construct alignment data on which a model learns to separate a word from
the request it appears in.

In the substitution experiments, most of the reduction in refusal comes
from replacing the most feared word at all: a random harm-free synonym
drawn from the same candidates already removes $4$ to $24$ points of
refusal. Choosing the synonym the table places lowest adds a smaller
increment of $2$ to $8$ points, positive in all eight models but
Holm-significant in only three of the seven confirmation models
(Table~\ref{tab:random-syn}). The table's own shift difference between a
word and its substitute also does not by itself rank the per-prompt
effects ($\rho=-0.05$ to $+0.16$); what predicts the effect is the shift
the replacement actually causes in the prompt. The table is therefore a
more reliable guide to \emph{which} word to replace than to \emph{which}
synonym to use.

On harmful prompts, the substitution brings a small rise in response
harmfulness.

\section*{Ethics Statement}

\paragraph{Purpose.}
This work aims to diagnose over-refusal: cases where a model declines a
benign request because of a single word. The vocabulary table is
intended as an auditing tool for model developers and evaluators, and our
main recommendation, that safety evaluations control for vocabulary, is
meant to make refusal measurements more reliable.

\paragraph{Dual use.}
The same lexical sensitivity also lowers refusal on harmful prompts. On
OR-Bench Toxic, replacing the most feared word lowers refusal and brings
a small rise in response harmfulness (at most $+0.04$ under
GuardReasoner-8B), and appending a courage word to harmful prompts also
lowers refusal. We report these results because they show that part of
current safety behavior depends on surface wording, which model
developers need to know in order to fix it. The effects are modest, the
models studied are small open-weight models, and the techniques involved
(rewording a prompt, adding a style word) are already widely known. We
do not release any harmful model responses.

\paragraph{Release.}
We release code, the prompts drawn from public benchmarks, aggregate
results and the full per-word tables for research use.

\paragraph{Offensive content.}
The paper and its artifacts contain offensive terms, including slurs and
sexual terms, because these words sit at the top of the tables. We list
them only where needed to report results, with a content warning in
Appendix~\ref{app:scan}.

\paragraph{Data and annotation.}
Prompts come from public datasets: OR-Bench, XSTest, WildJailbreak,
Dolly and No Robots, with StrongREJECT, IFEval and GSM8K as evaluation
endpoints. The harm-word lexicon follows the HarmBench and SORRY-Bench
taxonomies, and the vocabulary scan uses wordfreq and the CMU bad-words
list. All are used within their licenses. The human labels used to
validate GuardReasoner-8B are those released with XSTest.

\bibliography{references}

\appendix

\section{Crossed ANOVA and cross-fitting}
\label{app:derivations}

For prompt $p$, matched word pair $w$ and boundary position $t$, the
neutral-corrected shift $D_{pwt}=h^{H}_{pwt}-h^{N}_{pwt}$ decomposes as in
Equation~\ref{eq:anova}, per position, with
$\mu_t=\overline D_{\cdot\cdot t}$,
$A_{pt}=\overline D_{p\cdot t}-\mu_t$,
$B_{wt}=\overline D_{\cdot wt}-\mu_t$ and
$I_{pwt}=D_{pwt}-\mu_t-A_{pt}-B_{wt}$. On the balanced grid these are the
exact least-squares two-way vector ANOVA: the prompt and word effects sum
to zero, so averaging cancels both nuisance axes and isolates $\mu_t$. At
state 30 the four terms hold $0.30$, $0.32$, $0.15$ and $0.23$ of the
squared norm at \texttt{</think>} and $0.21$, $0.33$, $0.16$ and $0.30$
at the following token; the 24 word means are close to rank one
(rank-1 energy $0.76$--$0.89$); the three words below cosine $0.82$
with $\mu$ are the weapon words \emph{rifle}, \emph{bomb} and
\emph{grenade}, which are also the weakest behaviorally.

Prompts and words receive independent four-fold assignments. For test
fold $f$,
\begin{equation}
  \mu_{ft}=\frac{1}{|P_{-f}||W_{-f}|}
  \sum_{p:f_p\ne f}\sum_{w:f_w\ne f}D_{pwt},
  \label{eq:crossfit}
\end{equation}
with $f_p=f_w=f$ at test time. The held-out mean direction has cosine
$0.90$--$0.97$ with the train-fitted direction in every fold. For the
projection--behavior association, both the projection and the refusal
change are residualized by the prompt-plus-word fixed-effect design
matrix $X$, using $M_X=I-X(X^\top X)^+X^\top$.

\begin{figure}[t]
  \centering
  \includegraphics{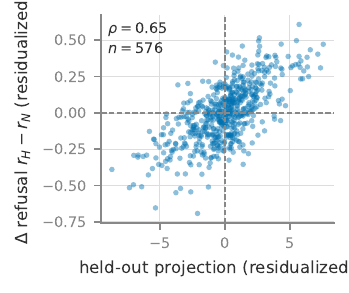}
  \caption{A held-out cell's projection
  $\langle D_{pw},\hat\mu_f\rangle$ (horizontal, residualized)
  predicts its refusal change (Spearman $\rho=0.65$ after
  residualizing both on prompt-plus-word fixed effects).}
  \label{fig:projection}
\end{figure}

The intervention writes the raw mean shift, not its unit vector, at the
two boundary positions. On a fold-$f$ held-out cell it subtracts
$a\mu_{ft}$ from the harm version and adds it to the neutral version,
\begin{equation}
  \begin{aligned}
  \widetilde h^H_{pwt}(a)&=h^H_{pwt}-a\mu_{ft},\\
  \widetilde h^N_{pwt}(a)&=h^N_{pwt}+a\mu_{ft},
  \end{aligned}
  \label{eq:write}
\end{equation}
so that, position by position, the boundary gap becomes
\begin{equation}
  \begin{aligned}
  \widetilde D_{pw}(a)&=(1-2a)\,\mu_f+R^{(f)}_{pw},\\
  R^{(f)}_{pw}&:=D_{pw}-\mu_f .
  \end{aligned}
  \label{eq:intervention}
\end{equation}
The dose $a^*=0.5$ exactly removes the fitted shared component and
leaves the fold-relative prompt, word and interaction remainder
$R^{(f)}$ untouched. It is also the dose a benign-KL ceiling admits:
over 256 sealed benign prompts the mean
$D_{\mathrm{KL}}(p_0\,\|\,p_a)$ between the base and written next-token
distributions at the boundary is $0.003$, $0.008$, $0.031$ and $0.128$
at magnitudes $0.125$, $0.25$, $0.5$ and $1$, and $a^*$ is the largest
magnitude whose KL stays monotone in dose and at most $0.10$ overall and
in every fold, at both signs. The two null directions of
Section~\ref{sec:causal}, a random direction orthogonal to $\mu_f$ and a
sign-permutation of $\mu_f$, were rescaled until they matched that KL
(multipliers $2.1$--$2.3$ and $0.84$). The write at $a^*$ moves the
boundary gap by $+0.083$ [$+0.053, +0.115$] of refusal in the intended
direction, against $-0.005$ for the random null and $+0.028$ for the
permutation null. Both margins are Holm-significant
($p\le2\times10^{-3}$). The effect is carried almost entirely by the
final boundary token ($0.070$ when it alone is written, against $0.011$
for the token before it, with no synergy), and adding the direction at
$a^*$ raises refusal on 256 sealed benign prompts by $+0.058$
[$+0.031, +0.085$].

\section{Dose unit and deep doses}
\label{app:dose}

The write adds $a\cdot\mu$ to the residual stream at the output of one
decoder block at the two boundary positions, once, during prefill; $\mu$
is the model's own consensus direction (the mean of its four fold
directions), so dose $1$ is the model's own mean harm-minus-neutral
boundary shift. The block is chosen by the fixed rule
$\mathrm{round}(29/33\cdot n_{\text{blocks}})$; direction norms range
from $0.90$ (Qwen3.5-0.8B) to $42.6$ (Qwen3.5-27B).

\begin{figure}[t]
  \centering
  \includegraphics[width=0.85\columnwidth]{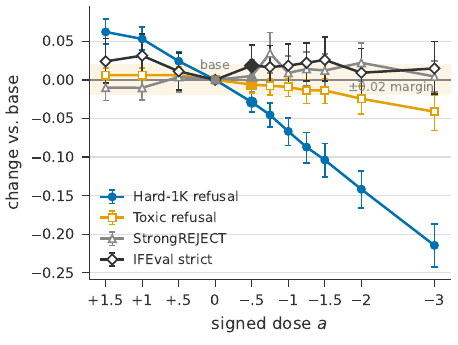}
  \caption{Dose response of the Qwen3.5-4B boundary write on four
  external endpoints, written minus base (base rates: Hard-1K $0.886$,
  Toxic $0.986$, StrongREJECT $0.063$, IFEval $0.804$). Zero is base by
  construction; positive doses are sign-flipped controls; filled markers
  are the registered dose $-0.5$. Bars are two-sided bootstrap intervals
  at the registered Bonferroni level ($98.3\%$ for Hard-1K, $99.4\%$ for
  the guards); the shaded band is the registered $\pm0.02$
  non-inferiority margin for the guard endpoints.}
  \label{fig:dose-write}
\end{figure}

\paragraph{External endpoints.}
On the frozen Qwen3.5-4B snapshot we wrote the consensus direction at ten
doses from $+1.5$ to $-3$ and measured four external benchmarks
(Figure~\ref{fig:dose-write}): OR-Bench Hard-1K (1{,}319 seemingly toxic
but benign prompts) for over-refusal, OR-Bench Toxic (655 prompts) for
the refusal the model should keep, StrongREJECT
\citep{souly2024strongreject} (313 prompts, finetuned evaluator) for the
harmfulness of what the model actually says, and IFEval
\citep{zhou2023ifeval} (541 prompts, strict prompt-level accuracy) for
instruction following. Refusal is judged by GuardReasoner-8B
(Appendix~\ref{app:judges}). The dose $-0.5$ was registered in a plan frozen before
generation, with Hard-1K as the superiority endpoint (refusal must fall
by at least $0.02$, one-sided $\alpha=0.05/3$) and the other three as
non-inferiority guards with an absolute $0.02$ margin against their
adverse directions (one-sided $\alpha=0.05/9$); deltas are prompt-paired
with 20{,}000-resample bootstraps and two-sided intervals at the
registered Bonferroni level ($98.3\%$ for Hard-1K, $99.4\%$ for the
guards).

At $-0.5$ the write lowers Hard-1K refusal by $-0.029$ [$-0.042, -0.017$]
from a base rate of $0.886$ and clears the superiority gate (adjusted
upper bound $-0.017$); Toxic moves by $-0.006$ [$-0.017, +0.003$],
StrongREJECT by $+0.005$ [$-0.015, +0.026$] and IFEval by $+0.018$
[$-0.007, +0.045$]. Non-inferiority holds formally on Toxic and IFEval;
on StrongREJECT the one-sided bound ($+0.024$) lands outside the
margin. The
sign-flipped dose $+0.5$ raises Hard-1K refusal by $+0.024$
[$+0.012, +0.036$], a mirror of the removal, and moves every other
endpoint by at most $0.011$: pushing the model up its ``fear'' axis
raises refusal of benign prompts and little else.

\paragraph{Deep doses.}
The nine further doses ran through the identical pipeline afterwards and
are reported descriptively. Hard-1K refusal falls monotonically from
$-0.029$ at $-0.5$ to $-0.215$ at $-3$, six times the registered dose $-0.5$, and
rises dose-ordered on the positive side ($+0.024$, $+0.053$, $+0.062$ at
$+0.5$, $+1.0$, $+1.5$). The cost that scales is Toxic refusal, sliding
from $-0.006$ to $-0.041$ at $-3$; StrongREJECT harmfulness stays within
$0.035$ of base, its interval excluding zero only at $-0.75$ ($+0.035$);
IFEval stays flat to positive and GSM8K \citep{cobbe2021gsm8k} accuracy
moves by at most $0.009$ in absolute value at any dose.

\section{Insertion and the direction across the eight models}
\label{app:insertion-xmodel}

\begin{figure*}[t]
  \centering
  \includegraphics{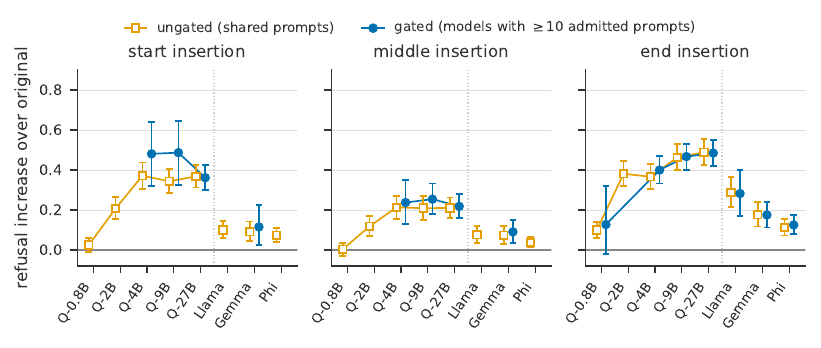}
  \caption{One inserted harm word raises refusal in all eight models.
  Paired increase over the regenerated original at each position, over
  all shared prompts (orange) and over gate-admitted prompts (blue, drawn
  where at least ten prompts are admitted), with $95\%$ bootstrap
  intervals.}
  \label{fig:insertion-xmodel}
\end{figure*}

\paragraph{Insertion.}
The cohort is the subset of the 321 insertion prompts that every one of
the eight models had answered in at least $20\%$ of ten earlier samples:
130 prompts. Each keeps its word, and the texts are identical across
models; each model runs its own detection gate. Intervals are $95\%$
two-stage paired bootstraps (10{,}000 resamples). 22 of the 24
model--position increases exclude zero after Holm correction
(Table~\ref{tab:insertion-xmodel}, Figure~\ref{fig:insertion-xmodel}).

\begin{table*}[t]
  \centering
  \caption{One inserted harm word across eight models. $r_O$: refusal of
  the regenerated original. Ungated: paired increase over all shared
  prompts with $95\%$ bootstrap interval. Gated: number of prompts
  admitted by the model's own detection gate, and the increase on them.
  $^\dagger$: Holm adjusted $p\ge0.05$. $^\ddagger$: $n=129$.}
  \label{tab:insertion-xmodel}
  \footnotesize
  \setlength{\tabcolsep}{2.4pt}
  \begin{tabular}{lccccccc}
    \toprule
    & & \multicolumn{3}{c}{Ungated increase} & \multicolumn{3}{c}{Gated: $n$ admitted, increase} \\
    \cmidrule(lr){3-5}\cmidrule(lr){6-8}
    Model & $r_O$ & Start & Middle & End & Start & Middle & End \\
    \midrule
    Qwen3.5-0.8B & $.10$ & $+.02$ [$-.01,+.06$]$^\dagger$ & $+.00$ [$-.03,+.03$]$^\dagger$ & $+.10$ [$+.06,+.14$] & 0, -- & 0, -- & 11, $+.13^\dagger$ \\
    Qwen3.5-2B & $.22$ & $+.21$ [$+.15,+.26$] & $+.12$ [$+.07,+.17$] & $+.38$ [$+.32,+.45$] & 0, -- & 0, -- & 6, $+.58$ \\
    Qwen3.5-4B & $.22$ & $+.37$ [$+.30,+.44$] & $+.21$ [$+.16,+.27$] & $+.37$ [$+.30,+.43$] & 21, $+.48$ & 43, $+.24$ & 111, $+.40$ \\
    Qwen3.5-9B & $.17$ & $+.34$ [$+.29,+.40$] & $+.21$ [$+.15,+.27$] & $+.46$ [$+.40,+.53$] & 23, $+.49$ & 83, $+.25$ & 125, $+.47$ \\
    Qwen3.5-27B$^\ddagger$ & $.22$ & $+.37$ [$+.31,+.43$] & $+.21$ [$+.16,+.26$] & $+.49$ [$+.42,+.56$] & 117, $+.36$ & 105, $+.22$ & 127, $+.49$ \\
    Llama-3.2-3B & $.03$ & $+.10$ [$+.06,+.15$] & $+.08$ [$+.04,+.12$] & $+.29$ [$+.21,+.36$] & 0, -- & 2, $-.05^\dagger$ & 57, $+.28$ \\
    Gemma-4-E2B & $.04$ & $+.09$ [$+.04,+.14$] & $+.07$ [$+.03,+.12$] & $+.18$ [$+.12,+.24$] & 39, $+.12$ & 95, $+.09$ & 125, $+.18$ \\
    Phi-4-mini & $.04$ & $+.07$ [$+.04,+.11$] & $+.04$ [$+.01,+.07$] & $+.11$ [$+.07,+.16$] & 0, -- & 3, $+.00^\dagger$ & 100, $+.12$ \\
    \bottomrule
  \end{tabular}
\end{table*}

\paragraph{The direction.}
Every value in Table~\ref{tab:direction-xmodel} is on the 136 held-out units
(408 cells per model, 64 distinct words;
Appendix~\ref{app:exclusion}). For the link, each of a
prompt's insertion texts gives one cell, and the inserted word's shift
along the direction is rank-correlated with the paired refusal increase
after position fixed effects. For the write, the direction is added at
doses $-0.75$ and $-1.5$ on the end-inserted text and $+0.75$ and $+1.5$
on the original, reusing the sampling seeds of the undosed arms. Two
controls are written at the same dose: random directions of the same
norm orthogonal to the model's direction, and one random direction whose
dose is raised until its benign KL matches the model's direction. $p$
values are Holm-adjusted over the seven confirmation models.

\begin{table*}[t]
  \centering
  \caption{Each model's own direction tracks and controls the inserted
  word's refusal rise, on 136 units.
  $\rho$: rank correlation of the projection with the refusal increase;
  partial: with the shift norm held fixed. Remove: refusal lowered on the
  end-inserted text at dose $-1.5$. Create: refusal raised on the
  original at $+1.5$. Own${}-{}$random, own${}-{}$KL: removal by the
  model's direction minus removal by the norm-matched and the KL-matched
  random directions. $95\%$ bootstrap intervals.}
  \label{tab:direction-xmodel}
  \footnotesize
  \setlength{\tabcolsep}{2.4pt}
  \begin{tabular}{lcccccc}
    \toprule
    & \multicolumn{2}{c}{Link} & \multicolumn{4}{c}{Write} \\
    \cmidrule(lr){2-3}\cmidrule(lr){4-7}
    Model & $\rho$ & Partial & Remove & Create & Own${}-{}$random & Own${}-{}$KL \\
    \midrule
    Qwen3.5-0.8B & $.32$ & $.22$ & $+.10$ [$+.06,+.13$] & $+.11$ [$+.07,+.15$] & $+.09$ [$+.05,+.12$] & $+.06$ [$+.03,+.08$] \\
    Qwen3.5-2B & $.63$ & $.31$ & $+.28$ [$+.23,+.32$] & $+.39$ [$+.34,+.43$] & $+.30$ [$+.26,+.35$] & $+.30$ [$+.25,+.35$] \\
    Qwen3.5-4B & $.64$ & $.42$ & $+.27$ [$+.22,+.32$] & $+.34$ [$+.29,+.39$] & $+.27$ [$+.22,+.32$] & $+.27$ [$+.22,+.32$] \\
    Qwen3.5-9B & $.77$ & $.53$ & $+.22$ [$+.18,+.27$] & $+.25$ [$+.20,+.30$] & $+.22$ [$+.18,+.26$] & $+.11$ [$+.06,+.15$] \\
    Qwen3.5-27B & $.69$ & $.46$ & $+.25$ [$+.20,+.30$] & $+.28$ [$+.22,+.33$] & $+.26$ [$+.21,+.31$] & $+.11$ [$+.07,+.15$] \\
    Llama-3.2-3B & $.54$ & $.32$ & $+.11$ [$+.07,+.15$] & $+.04$ [$+.02,+.07$] & $+.10$ [$+.07,+.14$] & $+.10$ [$+.06,+.14$] \\
    Gemma-4-E2B & $.50$ & $.34$ & $+.11$ [$+.07,+.16$] & $+.09$ [$+.05,+.14$] & $+.09$ [$+.06,+.13$] & $+.08$ [$+.05,+.13$] \\
    Phi-4-mini & $.34$ & $.26$ & $+.09$ [$+.05,+.13$] & $+.10$ [$+.07,+.14$] & $+.10$ [$+.07,+.14$] & $+.07$ [$+.04,+.10$] \\
    \bottomrule
  \end{tabular}
\end{table*}

\section{The vocabulary scan}
\label{app:scan}

\paragraph{Prompts and candidates.}
The Qwen3.5-4B table is scanned on 48 of the 96 over-refused OR-Bench
Hard-1K prompts of the courage-word test (mean 19 words; the scan ran
over that test's prompts in a fixed order and was stopped after 132 of
the 192, 48 of them over-refused). The clean-benign table is scanned on
40 clean benign prompts drawn from Dolly and No Robots, 90 to 170
words long, with no lexicon word in any inflected form and no
harm-adjacent stem; Qwen3.5-4B refuses $0.003$ of their samples. The
71{,}583 candidates are the 43{,}544 single-token vocabulary words of at
least three lowercase letters and the 28{,}027 multi-token words among
the 50k most frequent English words, plus the 12 insertion-lexicon
words and neutral twins that fall in neither set. Each word is appended as
\texttt{text + " " + w}; the prompt is prefilled once and the word plus
the template tail is fed from the cached state. The shift is the projection
change at the two \tbg{} positions, summed, in units of the Hard-1K
projection standard deviation.

\paragraph{Shape and stability.}
Percentiles of the mean shift in the Qwen3.5-4B table: min $-3.82$, p1
$-1.16$, p10 $-0.43$, median $-0.09$, p90 $+0.15$, p99 $+0.47$, p99.9
$+0.77$, max $+1.05$; $70\%$ of words fall below zero. Two disjoint
24-prompt halves rank the words at $\rho=+0.95$ ($+0.91$ to $+0.97$
over 200 random splits) and share 81 of their top 100 on average. The
99-word insertion lexicon averages $+0.49$ sd; all 99 words lie above
the 84th percentile and 52 in the top $1\%$ (\emph{shoot} 91st
percentile, \emph{bomb} 97th, \emph{gun} 98th, \emph{meth} 98th,
\emph{hack} 99th). The clean-benign table sits higher and is
compressed at the bottom (p1 $-0.09$, median $+0.13$, p99 $+0.66$, max
$+2.52$; $4.6\%$ of words below zero; its two disjoint 20-prompt halves
agree at $\rho=+0.93$), and in it the lexicon's weapon, cyber and drug
words sit mid-distribution (\emph{shoot} 25th percentile, \emph{meth}
42nd, \emph{hack} 49th, \emph{gun} 65th, \emph{bomb} 70th).

\paragraph{Which prompts to scan.}
On Qwen3.5-4B we compared four tables over the same 71{,}583 words,
scanned on the 40 clean benign prompts, on 48 over-refused Hard-1K
prompts (BP), on 84 harmful prompts (HP), and on all of them. On the 64
held-out prompts that chose the table, the BP table ranks the 40 test
words of Section~\ref{sec:val-insert} at $\rho=+0.90$ [$+0.86, +0.92$],
the HP table at $+0.88$, the mixed table at $+0.88$ and the clean-benign
table at $+0.57$; on 64 further prompts drawn by the same rule after the
choice, excluding every prompt any table or earlier test had used (the
Qwen3.5-4B prompts of Section~\ref{sec:val-insert}), the order holds: BP
$+0.92$ [$+0.89, +0.94$], mixed $+0.89$, HP $+0.88$, clean-benign $+0.55$,
BP leading on the benign prompts and HP on the harmful ones. Over the
47{,}371 frequent words the clean-benign and BP tables correlate at
Spearman $+0.32$. On the 24 words where the tables agree they all reach
$+0.91$ to $+0.96$; the clean-benign table loses on words high in it alone
(\emph{proteins}, \emph{phytoplankton}), which leave refusal unchanged.
The seven other models therefore scan their tables on the eight BP
prompts whose own ranking agrees best with the BP table.

\paragraph{The feared words.}
\emph{Content warning: offensive terms.} The ten most feared distinct
words of the Qwen3.5-4B table (after the same drop rules) are
\emph{pedophile}, \emph{bestiality}, \emph{gangbang}, \emph{rapist},
\emph{creampie}, \emph{smut}, \emph{infanticide}, \emph{erotic}, the
singular n-word and \emph{blowjob} (shift $+0.94$ to $+1.05$ sd); one,
\emph{pedophile}, is a lexicon word. The ten words tested in
Section~\ref{sec:table}, the most feared distinct words of the
clean-benign table on its 20 selection prompts, are \emph{pedophilia}, the
plural n-word, \emph{masturbate}, \emph{blowjobs}, \emph{rapist},
\emph{pornographic}, \emph{bestiality}, \emph{antisemitic},
\emph{penises} and \emph{hentai} (shift there $+1.44$ to $+2.11$ sd;
percentile $99.8$ to $100$ in the Qwen3.5-4B table). None is a
lexicon word. On 20 held-out prompts they give refusal rates of $0.255$
to $0.575$ against $0.000$--$0.020$ for their twins (one twin,
\emph{watercolor}, $0.090$).

\begin{table}[t]
  \centering
  \footnotesize
  \setlength{\tabcolsep}{3pt}
  \caption{Words selected from the two tails of the clean-benign table;
  pct: percentile in the Qwen3.5-4B table. Top: refusal rates for the ten
  feared words on 20 held-out clean benign prompts (exact words listed
  in this appendix). Bottom: refusal changes for the ten courage
  words against neutral twins on 96 harmful
  (HP) and 96 over-refused benign (BP) prompts, negative values being
  refusal removed; all Holm $p\le0.008$.}
  \label{tab:words}
  \begin{tabular}{lccc}
    \toprule
    feared words & pct & \multicolumn{2}{c}{refusal} \\
    \midrule
    top ten (sexual terms, & $99.8$ to & \multicolumn{2}{c}{$0.26$ to} \\
    \quad slurs, pedophilia) & $100$ & \multicolumn{2}{c}{$0.58$} \\
    \midrule
    courage word & pct & HP & BP \\
    \midrule
    poetic & $0.00$ & $-0.361$ & $-0.452$ \\
    whimsical & $0.01$ & $-0.319$ & $-0.456$ \\
    playfully & $0.02$ & $-0.192$ & $-0.348$ \\
    lyrical & $0.36$ & $-0.192$ & $-0.076$ \\
    flowery & $0.46$ & $-0.176$ & $-0.176$ \\
    sarcastically & $0.01$ & $-0.173$ & $-0.692$ \\
    suspenseful & $0.11$ & $-0.159$ & $-0.333$ \\
    cartoonish & $0.19$ & $-0.105$ & $-0.157$ \\
    lighthearted & $0.21$ & $-0.105$ & $-0.240$ \\
    creatively & $3.58$ & $-0.080$ & $-0.070$ \\
    \bottomrule
  \end{tabular}
\end{table}

\begin{table}[t]
  \centering
  \footnotesize
  \setlength{\tabcolsep}{2.5pt}
  \caption{Held-out insertion dose-response per model, pooled over 16
  benign and 16 harmful held-out prompts (32 and 32 on Qwen3.5-4B).
  Top$-$bottom: refusal change of the top bin minus the bottom bin with
  $95\%$ interval. In order: adjacent bin pairs in order, of 7. $\rho$:
  Spearman of the table shift with the refusal change over the 40 words
  (39 on Llama), own table and the Qwen3.5-4B table.}
  \label{tab:xmodel-table}
  \setlength{\tabcolsep}{2pt}
  \begin{tabular}{lcccc}
    \toprule
    model & top$-$bottom & in order & $\rho$ own & $\rho$ Qwen \\
    \midrule
    Qwen3.5-0.8B & $+.71$ [$+.67, +.75$] & 7 & $+.96$ & $+.74$ \\
    Qwen3.5-2B & $+.39$ [$+.31, +.47$] & 6 & $+.82$ & $+.76$ \\
    Qwen3.5-4B & $+.53$ [$+.48, +.58$] & 7 & $+.92$ & -- \\
    Qwen3.5-9B & $+.50$ [$+.44, +.57$] & 7 & $+.83$ & $+.85$ \\
    Qwen3.5-27B & $+.32$ [$+.24, +.41$] & 5 & $+.78$ & $+.70$ \\
    Llama-3.2-3B & $+.44$ [$+.33, +.56$] & 6 & $+.54$ & $+.59$ \\
    Gemma-4-E2B & $+.39$ [$+.28, +.50$] & 5 & $+.78$ & $+.71$ \\
    Phi-4-mini & $+.57$ [$+.51, +.63$] & 7 & $+.91$ & $+.72$ \\
    \bottomrule
  \end{tabular}
\end{table}

\paragraph{What predicts a word's shift.}
Table~\ref{tab:wordrep} gives the out-of-fold accuracy of ridge
regressions from context-free representations of the word to its shift
in the clean-benign table, with folds grouped by stem.

\begin{table}[t]
  \centering
  \footnotesize
  \setlength{\tabcolsep}{3pt}
  \caption{Predicting a word's shift in the clean-benign table from its
  context-free representation on Qwen3.5-4B (71{,}583 words, stem-grouped
  cross-validation).}
  \label{tab:wordrep}
  \begin{tabular}{lcc}
    \toprule
    representation & Spearman & $R^2$ \\
    \midrule
    token count, frequency, length & $+0.46$ & $0.17$ \\
    input embedding & $+0.78$ & $0.53$ \\
    value vectors, layers 23--31 & $+0.81$ to $+0.83$ & $0.65$--$0.68$ \\
    word-alone residual, blocks 7--23 & $+0.83$ to $+0.84$ & $0.70$--$0.72$ \\
    word-alone chat, boundary states & $+0.87$ & $0.80$ \\
    \midrule
    projection on the direction & $-0.35$ to $+0.04$ & -- \\
    \bottomrule
  \end{tabular}
\end{table}

\section{Attention reads and MLP writes on Qwen3.5-4B}
\label{app:mechanism}

Qwen3.5-4B has 32 blocks with 24 Gated DeltaNet mixers and 8
softmax-attention layers. On the crossed grid, the harm word receives
the same attention as its neutral twin at the end-of-turn token (summed
over all 128 heads $7.17$ against $7.36$, $\Delta=-0.19$
[$-0.61, +0.28$]), and given the same query the harm word's key is less
attractive ($-0.39$ logits). Of the harm-minus-neutral shift along the
direction, the 30 MLPs write $73\%$, the DeltaNet mixers $24\%$ and the
softmax-attention mixers $3\%$, nearly all from layer 18 on; the direct
reads of the word by the softmax heads account for $1\%$. No attention
statistic predicts the refusal change after prompt and word fixed effects
($|\rho|\le0.13$), while the boundary projection does ($\rho=+0.65$).
What the heads read is still a harm feature: value vectors at the word
token separate unseen harm from neutral words at $0.92$--$0.98$
accuracy.

\section{Substitution texts}
\label{app:subst}

\begin{figure*}[t]
  \centering
  \includegraphics[width=\textwidth]{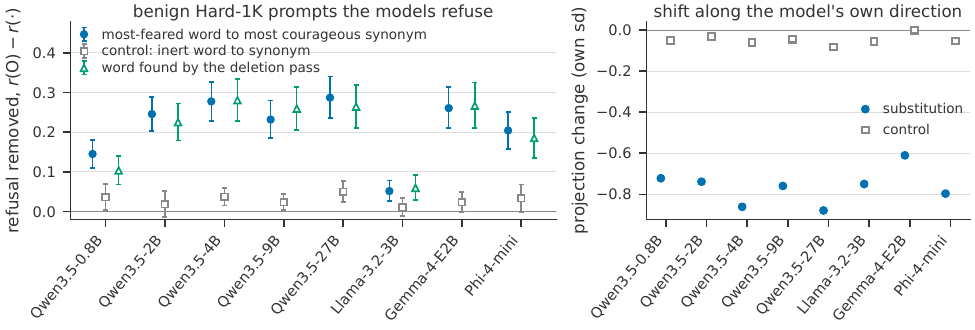}
  \caption{Table-guided in-place substitution on the benign Hard-1K
  prompts the models refuse ($95\%$
  two-stage paired bootstrap intervals). Left: refusal removed by
  replacing the most feared word by its most courageous synonym (filled),
  by replacing the prompt's most inert word by a synonym (squares), and,
  for reference, by replacing the word found by the direction's deletion
  pass (triangles). Right: the change in the boundary projection on the
  model's own direction, in that model's Hard-1K standard deviation.}
  \label{fig:subst}
\end{figure*}

\paragraph{Table nomination.}
Every whitespace word of a prompt, edge punctuation stripped and
lower-cased, is looked up in the model's table (exact, then inflection
fallback, then for a hyphenated word its highest-shift part); stop words
are excluded. The nominee is the highest-shift content word (median
within-prompt $z$ $2.4$--$3.4$, median percentile $92$--$96$). The drafter
returns up to ten substitutes (about $7$ pass the checks); the accepted
candidate with the lowest table shift is used, which is the drafter's
own first choice only $8$--$12\%$ of the time. Hedge units (``what
steps'', ``without getting caught'') are replaced as phrases. The
structural checks reject a substitute that contains a lexicon word or
another prompt's nominee, that equals the nominee, or that creates a
repeated adjacent word. The control replaces the content word whose
table shift is nearest zero within a frequency band by the drafter's
synonym.

\paragraph{Deletion-pass nomination.}
The harmful side and the reference arm of Figure~\ref{fig:subst} locate
the word with the direction instead: each word, and each adjacent pair
containing a content word, is deleted in turn, the prompt is re-read,
and the unit whose deletion lowers the projection most is nominated. On
the harmful prompts the nominated unit is sharper (median $z$ $3.4$ to
$4.2$ against $2.5$ to $3.5$ on benign prompts), and on XSTest the safe
and contrast prompts share the top word \emph{kill} yet project apart at
AUC $0.94$ to $0.998$: the direction reads the word in its context.

\paragraph{An independent meaning check.}
Whether an edited prompt still asks for what the original asked for was
read once more by a model with no part in the drafting (GPT-6 Luna, the
lowest reasoning effort the route accepts, temperature $0$, fixed seed,
structured output), on all $4{,}805$ distinct (original, edited) pairs of
the three substitution runs ($7{,}713$ uses across the eight models). It
reads $0.61$ of the shared-text substitutions, $0.58$ of the harmful ones
and $0.33$ of the table-guided ones as the same request, against $0.86$
and $0.81$ for the two inert-word controls and $0.07$ to $0.13$ for the
hedge removals; a second seed on a tenth of the pairs reproduces its
verdict on $0.945$ of them (Cohen's $\kappa=0.89$). On the pairs it
accepts, replacing the nominated word still
removes $+0.02$ to $+0.16$ of refusal in the shared-text run, $+0.05$ to
$+0.19$ with the table nomination and $+0.01$ to $+0.11$ on the harmful
prompts, $21$ of those $24$ intervals excluding zero, while the
inert-word control stays at $+0.01$ to $+0.04$.

The same pairs, prompt and schema then went to a second judge from a
third family (Claude Haiku 5.5, again the lowest reasoning effort the
route accepts; that route offers neither temperature nor seed, so its
repeat is a second call rather than a second seed, and it returns no
label for two pairs). It is the more permissive reader: it accepts
$0.79$ of the shared-text substitutions, $0.71$ of the harmful ones and
$0.53$ of the table-guided ones, against $0.97$ and $0.95$ for the
controls and $0.09$ to $0.12$ for the hedge removals, and repeats its own
verdict on $0.992$ of the sampled pairs (Cohen's $\kappa=0.98$). The two
judges put the arms in the same order and agree on $0.800$ of the
$4{,}803$ pairs both labelled ($\kappa=0.58$), disagreeing almost only in
one direction: the second accepts $882$ pairs the first rejects and
rejects $81$ the first accepts. On the $2{,}588$ pairs both judges accept,
replacing the nominated word removes $+0.02$ to $+0.16$, $+0.05$ to
$+0.20$ and $+0.01$ to $+0.10$ of refusal in the three runs, $19$ of those
$24$ intervals excluding zero, and on the second judge's own larger
subset all $24$ exclude zero; the inert-word control stays within
$\pm0.04$ throughout.

\begin{table*}[t]
  \centering
  \caption{Table-guided substitution on the benign Hard-1K prompts
  ($95\%$ two-stage paired bootstrap
  intervals). $n$: prompts with a substitute. Deletion pass: the
  substitution of Section~\ref{sec:natural-phen} rerun with the word
  found by the direction's deletion pass on this cohort (prompts most
  models refuse, fitting cohort excluded), not on the 476-prompt
  shared-text cohort of Table~\ref{tab:natural-benign}. Projection: change of the boundary projection in
  the model's Hard-1K sd. $\rho$: per-prompt Spearman between projection
  change and refusal change. XSTest: refusal removed on the safe XSTest
  prompts.}
  \label{tab:subst}
  \footnotesize
  \setlength{\tabcolsep}{2.6pt}
  \resizebox{\textwidth}{!}{%
  \begin{tabular}{lrcccccccc}
    \toprule
    model & $n$ & $r(\textsc{O})$ & $r(\textsc{O})-r(\textsc{S})$ & $r(\textsc{O})-r(\textsc{C})$ & deletion pass & projection S / C & $\rho$ & XSTest S \\
    \midrule
    Qwen3.5-0.8B & 246 & .331 & $+.145$ [$+.110, +.181$] & $+.036$ & $+.104$ & $-0.72$ / $-0.05$ & $+0.54$ & $+.060$ [$+.022, +.102$] \\
    Qwen3.5-2B & 249 & .629 & $+.246$ [$+.203, +.288$] & $+.018$ & $+.225$ & $-0.74$ / $-0.03$ & $+0.66$ & $+.082$ [$+.035, +.134$] \\
    Qwen3.5-4B & 251 & .929 & $+.278$ [$+.229, +.326$] & $+.037$ & $+.281$ & $-0.86$ / $-0.06$ & $+0.80$ & $+.058$ [$+.014, +.109$] \\
    Qwen3.5-9B & 247 & .937 & $+.232$ [$+.185, +.280$] & $+.023$ & $+.259$ & $-0.76$ / $-0.04$ & $+0.76$ & $+.103$ [$+.052, +.164$] \\
    Qwen3.5-27B & 249 & .897 & $+.287$ [$+.236, +.340$] & $+.050$ & $+.265$ & $-0.88$ / $-0.08$ & $+0.79$ & $+.056$ [$+.011, +.107$] \\
    Llama-3.2-3B & 252 & .127 & $+.052$ [$+.027, +.078$] & $+.011$ & $+.060$ & $-0.75$ / $-0.06$ & $+0.31$ & $+.006$ [$+.000, +.018$] \\
    Gemma-4-E2B & 250 & .879 & $+.261$ [$+.210, +.314$] & $+.023$ & $+.267$ & $-0.61$ / $-0.00$ & $+0.71$ & $+.131$ [$+.059, +.210$] \\
    Phi-4-mini & 249 & .481 & $+.204$ [$+.158, +.251$] & $+.033$ & $+.186$ & $-0.80$ / $-0.05$ & $+0.75$ & $+.112$ [$+.053, +.176$] \\
    \bottomrule
  \end{tabular}}
\end{table*}

\begin{table}[t]
  \centering
  \footnotesize
  \setlength{\tabcolsep}{2.5pt}
  \caption{Table-lowest against random synonym on the benign Hard-1K
  prompts. $\Delta_S$: refusal removed by the table's lowest candidate;
  $\Delta_{\mathrm{rand}}$: by a random admissible candidate of the same
  unit; gap: $\Delta_S-\Delta_{\mathrm{rand}}$ with $95\%$ paired
  bootstrap interval; $^*$: one-sided Holm $p<0.05$ over the seven
  transfer models. Every model's random candidate also ends $0.12$ to
  $0.21$ sd higher on the direction.}
  \label{tab:random-syn}
  \resizebox{\columnwidth}{!}{%
  \begin{tabular}{lrccc}
    \toprule
    model & $n$ & $\Delta_S$ & $\Delta_{\mathrm{rand}}$ & gap \\
    \midrule
    Qwen3.5-0.8B & 241 & $+.145$ & $+.120$ & $+.025$ [$-.005, +.055$] \\
    Qwen3.5-2B & 247 & $+.246$ & $+.199$ & $+.047$ [$+.009, +.085$]$^*$ \\
    Qwen3.5-4B & 249 & $+.276$ & $+.243$ & $+.033$ [$-.004, +.072$] \\
    Qwen3.5-9B & 243 & $+.233$ & $+.208$ & $+.025$ [$-.011, +.061$] \\
    Qwen3.5-27B & 246 & $+.283$ & $+.216$ & $+.067$ [$+.030, +.107$]$^*$ \\
    Llama-3.2-3B & 248 & $+.052$ & $+.035$ & $+.017$ [$-.006, +.042$] \\
    Gemma-4-E2B & 248 & $+.265$ & $+.187$ & $+.078$ [$+.038, +.119$]$^*$ \\
    Phi-4-mini & 248 & $+.206$ & $+.170$ & $+.036$ [$+.004, +.069$] \\
    \bottomrule
  \end{tabular}}
\end{table}

\begin{table*}[t]
  \caption{The deletion-pass substitution on the benign Hard-1K prompts
  the models refuse (Section~\ref{sec:natural-phen}): refusal removed by
  the substitution (S), the inert control (C) and the hedge removal (R)
  under the explicit-refusal judge and under GuardReasoner-8B's refusal
  field, with $95\%$ paired bootstrap intervals. $n$: Hard-1K prompts the
  model refuses at base with the request read as unharmful.}
  \label{tab:natural-benign}
  \centering
  \scriptsize
  \setlength{\tabcolsep}{1.5pt}
  \resizebox{\textwidth}{!}{%
  \begin{tabular}{lrcccccc}
    \toprule
    & & \multicolumn{3}{c}{explicit-refusal judge} & \multicolumn{3}{c}{GuardReasoner refusal} \\
    \cmidrule(lr){3-5} \cmidrule(lr){6-8}
    model & $n$ & S & C & R & S & C & R \\
    \midrule
    Qwen3.5-0.8B & 239 & $+.079$ [$+.055, +.103$] & $-.001$ [$-.022, +.020$] & $+.073$ [$+.015, +.135$] & $+.111$ [$+.085, +.137$] & $+.009$ [$-.014, +.032$] & $+.123$ [$+.052, +.196$] \\
    Qwen3.5-2B & 360 & $+.170$ [$+.139, +.201$] & $+.008$ [$-.015, +.031$] & $+.246$ [$+.158, +.336$] & $+.202$ [$+.167, +.236$] & $+.008$ [$-.015, +.031$] & $+.346$ [$+.246, +.448$] \\
    Qwen3.5-4B & 476 & $+.277$ [$+.236, +.318$] & $+.007$ [$-.012, +.026$] & $+.504$ [$+.384, +.619$] & $+.286$ [$+.246, +.327$] & $+.014$ [$-.002, +.031$] & $+.555$ [$+.437, +.670$] \\
    Qwen3.5-9B & 519 & $+.218$ [$+.179, +.258$] & $+.005$ [$-.011, +.021$] & $+.448$ [$+.316, +.579$] & $+.224$ [$+.186, +.265$] & $+.005$ [$-.009, +.020$] & $+.484$ [$+.350, +.614$] \\
    Qwen3.5-27B & 495 & $+.220$ [$+.180, +.261$] & $+.019$ [$-.002, +.041$] & $+.405$ [$+.264, +.548$] & $+.232$ [$+.192, +.271$] & $+.009$ [$-.007, +.025$] & $+.484$ [$+.356, +.608$] \\
    Llama-3.2-3B & 33 & $+.033$ [$+.015, +.052$] & $+.005$ [$-.010, +.019$] & $+.039$ [$-.012, +.098$] & $+.028$ [$+.011, +.046$] & $+.005$ [$-.009, +.019$] & $+.039$ [$-.013, +.097$] \\
    Gemma-4-E2B & 340 & $+.244$ [$+.202, +.285$] & $-.005$ [$-.025, +.015$] & $+.364$ [$+.229, +.496$] & $+.231$ [$+.190, +.273$] & $-.006$ [$-.025, +.014$] & $+.361$ [$+.220, +.498$] \\
    Phi-4-mini & 163 & $+.143$ [$+.109, +.176$] & $+.003$ [$-.021, +.026$] & $+.130$ [$+.018, +.241$] & $+.195$ [$+.158, +.232$] & $+.007$ [$-.017, +.031$] & $+.180$ [$+.053, +.308$] \\
    \bottomrule
  \end{tabular}}
\end{table*}

\begin{table*}[t]
  \caption{Harmful OR-Bench Toxic prompts and the XSTest contrast
  prompts: refusal removed by the substitution (S) and the hedge removal
  (R) under the explicit-refusal judge and under GuardReasoner-8B's
  refusal field, the substitution effect on the contrast prompts, and
  the response harm GuardReasoner adds, with $95\%$ paired bootstrap
  intervals. $n$: Toxic prompts the model refuses at base with the
  request read as harmful; $h(\textsc{O})$: share of original responses
  read as harmful on the prompts with a substitute.}
  \label{tab:natural-harmful}
  \centering
  \scriptsize
  \setlength{\tabcolsep}{2.5pt}
  \resizebox{\textwidth}{!}{%
  \begin{tabular}{lrcccccc}
    \toprule
    & & \multicolumn{3}{c}{explicit-refusal judge} & GR refusal & \multicolumn{2}{c}{GuardReasoner harm} \\
    \cmidrule(lr){3-5} \cmidrule(lr){6-6} \cmidrule(lr){7-8}
    model & $n$ & S & R & S, contrast & S & $h(\textsc{O})$ & $h(\textsc{S})-h(\textsc{O})$ \\
    \midrule
    Qwen3.5-0.8B & 445 & $+.157$ [$+.123, +.191$] & $+.226$ [$+.076, +.383$] & $+.164$ [$+.054, +.278$] & $+.180$ [$+.145, +.216$] & .241 & $-.014$ [$-.043, +.015$] \\
    Qwen3.5-2B & 596 & $+.195$ [$+.158, +.231$] & $+.357$ [$+.195, +.518$] & $+.178$ [$+.075, +.283$] & $+.187$ [$+.151, +.222$] & .038 & $+.026$ [$+.009, +.043$] \\
    Qwen3.5-4B & 619 & $+.114$ [$+.084, +.143$] & $+.278$ [$+.128, +.442$] & $+.242$ [$+.138, +.356$] & $+.098$ [$+.072, +.126$] & .002 & $+.010$ [$+.004, +.018$] \\
    Qwen3.5-9B & 620 & $+.091$ [$+.066, +.118$] & $+.174$ [$+.044, +.333$] & $+.214$ [$+.122, +.315$] & $+.079$ [$+.056, +.105$] & .001 & $+.002$ [$-.000, +.004$] \\
    Qwen3.5-27B & 623 & $+.076$ [$+.050, +.103$] & $+.183$ [$+.044, +.350$] & $+.242$ [$+.135, +.358$] & $+.073$ [$+.049, +.098$] & .000 & $+.004$ [$-.000, +.010$] \\
    Llama-3.2-3B & 474 & $+.226$ [$+.189, +.265$] & $+.343$ [$+.171, +.529$] & $+.308$ [$+.187, +.437$] & $+.233$ [$+.195, +.272$] & .101 & $+.041$ [$+.018, +.064$] \\
    Gemma-4-E2B & 615 & $+.124$ [$+.088, +.160$] & $+.187$ [$+.000, +.395$] & $+.214$ [$+.097, +.339$] & $+.137$ [$+.106, +.171$] & .011 & $+.021$ [$+.008, +.035$] \\
    Phi-4-mini & 626 & $+.174$ [$+.140, +.210$] & $+.239$ [$+.076, +.414$] & $+.258$ [$+.155, +.361$] & $+.135$ [$+.104, +.167$] & .007 & $+.012$ [$+.004, +.022$] \\
    \bottomrule
  \end{tabular}}
\end{table*}

\section{The harmful prompt pool}
\label{app:pool}

The frozen harmful pool holds 13{,}527 prompts labelled harmful by
WildJailbreak \citep{jiang2024wildteaming}, drawn from its adversarial
subset. From 161{,}430 such prompts, an iterative adversarial filter kept
in each round the half that a five-fold out-of-fold TF-IDF logistic
regression separated least well from WildJailbreak's adversarial benign
prompts, so that a lexical classifier cannot tell the two sides apart;
three batches were drawn this way and deduplicated by clustering into a
pool of 15{,}000. Three LLM judges then rated each request's
feasibility, and a prompt was kept when at least two agreed it was
feasible (1{,}398 dropped by majority, 75 unstable). The harmful label is
the source dataset's, not a per-prompt judgment. A fixed split of the
retained prompts (seed 42) gives the 1{,}353-prompt evaluation set from
which every harmful cohort of this paper, including the 321 prompts of
the insertion experiment, is drawn.

\section{Sampling contract}
\label{app:sampling}

Every response is sampled with temperature $1.0$, top-$p$ $0.9$ and
top-$k$ $50$, ten sequences per prompt (twenty in the table
dose-response tests), up to 2{,}048 new tokens with one retry at 4{,}096
when a response is truncated. The seed is fixed per prompt and
condition, so that a condition's responses are regenerated identically
and the dosed arms reuse the seeds of the undosed ones. Prompts are
rendered as a single user turn with the model's chat template and the
generation prompt appended; on the Qwen3.5 models thinking is disabled,
which leaves an empty think block in the template.

\section{Judges}
\label{app:judges}

\paragraph{Explicit refusal.}
The refusal label is a single surface fact about a response: whether it
contains a sentence in which the assistant itself says it cannot, will
not, or is not allowed to do what was asked. Warnings, disclaimers and
silent redirects do not count; a refusal statement followed by delivered
content still counts. Every label comes from one pass of
DeepSeek-V4-Flash with this fixed instruction, reading the ten responses
of a prompt together. Three judge families (GPT-5.6, DeepSeek-V4-Flash,
GLM-5.2), each run twice, agree on two response sets of Qwen3.5-4B
(Table~\ref{tab:refusal-judge}); the agreement was not measured on the
other seven models' responses. Reading a prompt's ten responses in one
call can correlate their label errors within the prompt; every interval
in the paper is a prompt-paired bootstrap, which absorbs within-prompt
correlation.

\begin{table}[t]
  \centering
  \footnotesize
  \setlength{\tabcolsep}{3pt}
  \caption{Agreement on the explicit-refusal label across three judge
  families, two passes each. Unanimous: share of responses on which all
  six passes agree.}
  \label{tab:refusal-judge}
  \begin{tabular}{lccc}
    \toprule
    set & retest $\kappa$ & cross-judge $\kappa$ & unanimous \\
    \midrule
    250 responses & $0.968$--$1.000$ & $0.976$--$0.988$ & $98.0\%$ \\
    2{,}992 responses & $0.985$--$0.990$ & $0.975$--$0.978$ & -- \\
    \bottomrule
  \end{tabular}
\end{table}

\paragraph{GuardReasoner-8B.}
The second instrument, used for OR-Bench refusal in the deep-dose
measurements and for response harmfulness on the harmful prompts, is
GuardReasoner-8B \citep{liu2025guardreasoner} in one greedy pass, whose
refusal field and response-harmfulness field are read. Of seven
instruments compared on the 2{,}250 human-labelled XSTest completions it
agrees best with the human label ($95.1\%$, $\kappa=0.901$, sensitivity
$0.959$, specificity $0.942$); the two human annotators agree on
$97.9\%$.

\paragraph{Hijack.}
Whether a response treats the appended word as part of the request, its
topic or its style is judged response by response by the same
DeepSeek-V4-Flash model under a fixed yes/no instruction.

\section{The fitting cohort and full-cohort results}
\label{app:exclusion}

The direction of every model is fitted on 96 prompts crossed with 24 harm
and 24 neutral words (Section~\ref{sec:direction}). Every validation
excludes a unit whose prompt is one of the 96, or whose inserted,
nominated, substituted or control word is one of the 48. This removes 66
of the 130 cross-model insertion units (Section~\ref{sec:causal} uses the
remaining 64 together with 72 further units whose prompts lie outside
the 321 insertion prompts and the fitting cohort and whose words come from
the insertion lexicon minus the 24 harm words, generated and judged in
all eight models exactly as the original 130, for 136 units), 15 of the 96 harmful prompts of the courage-word test (refilled to 96 by the same rule), 9
of the 32 harmful held-out prompts on Qwen3.5-4B (refilled to 32 by the
same rule) and 13 harmful prompts
and one word over four of the other models in
Section~\ref{sec:val-insert} (each of those four harmful strata was
refilled to 16 by the same selection rule, so each model again has 32
prompts), and 45 to 48 of the 397
substitution units per model in Section~\ref{sec:val-subst}. The
analyses on the full cohorts reach the same conclusions: the cross-model
link gives $\rho=0.31$ to $0.75$, removal $+0.09$ to $+0.29$ and creation
$+0.05$ to $+0.36$; the courage words lower refusal on the original 96 harmful
prompts by $0.170$, every word Holm-significant; the cross-model
dose-response gives top minus bottom $+0.37$ to $+0.71$ with the own
table ahead on the same five models; and no model's substitution effect
moves by more than $0.017$.

\end{document}